\documentclass{article}

\usepackage{microtype}
\usepackage{graphicx}
\usepackage{subcaption}
\usepackage{booktabs} 
\usepackage{amsfonts}
\usepackage{amsmath}
\usepackage{nicefrac}
\usepackage{microtype}
\usepackage{xcolor}
\usepackage{graphicx}
\usepackage{multirow}
\usepackage{natbib}
\usepackage{comment}
\usepackage{caption}
\usepackage{tcolorbox}
\tcbset{
  promptbox/.style={
    colback=gray!5,
    colframe=gray!60,
    boxrule=0.5pt,
    arc=2pt,
    left=6pt,
    right=6pt,
    top=6pt,
    bottom=6pt
  }
}
\usepackage{hyperref}

\usepackage[preprint]{icml2026}

\usepackage{amsmath}
\usepackage{amssymb}
\usepackage{mathtools}
\usepackage{amsthm}

\usepackage[capitalize,noabbrev]{cleveref}

\theoremstyle{plain}

\theoremstyle{definition}

\theoremstyle{remark}

\usepackage[textsize=tiny]{todonotes}

\icmltitlerunning{Negative Before Positive: Asymmetric Valence Processing
in Large Language Models}

\begin{document}

\twocolumn[
  \icmltitle{Negative Before Positive: Asymmetric Valence Processing\\
in Large Language Models}



  \icmlsetsymbol{equal}{*}

  \begin{icmlauthorlist}
  \icmlauthor{Sohan Venkatesh}{mahe}
  \end{icmlauthorlist}
    
   \icmlaffiliation{mahe}{Manipal Institute of Technology Bengaluru}
    
   \icmlcorrespondingauthor{Sohan Venkatesh}{soh.venkatesh@gmail.com}
  \icmlkeywords{Valence Circuits, LLMs, Mechanistic Interpretability}

  \vskip 0.3in
]



\printAffiliationsAndNotice{}  

\begin{abstract}
Mechanistic interpretability has revealed how concepts are
encoded in large language models (LLMs) but emotional content
remains poorly understood at the mechanistic level.
We study whether LLMs process emotional valence through dedicated
internal structure or through surface token matching.
Using activation patching and steering on open-source LLMs, we
find that negative and positive valence are processed at different
network depths. Negative outcomes localize to early layers while positive outcomes peak at mid-to-late layers.
Holding topic fixed while flipping valence produces sign-opposite
responses, ruling out topic detection.
Steering with the good-news direction at the identified layers shifts neutral prompts toward positive valence, showing these layers encode valence as a manipulable direction.
Emotional valence in LLMs is localized, causal and steerable, making it a concrete target for interpretability-based oversight.
\end{abstract}

\section{Introduction}
\label{sec:intro}
 
Understanding what LLMs represent internally has become a central
goal of mechanistic
interpretability~\citep{elhage2021mathematical, olsson2022context}.
Prior work has shown that LLMs encode factual associations in
specific layers~\citep{meng2022locating}, implement recognizable
algorithms~\citep{wang2022interpretability} and represent abstract
concepts as linear directions in the residual
stream~\citep{tigges2023linear, park2023linear}.
Emotional content, however, has been studied primarily through probing and behavioral analysis rather than causal intervention. This matters: if models have internal valence representations that causally shape their outputs, those representations are directly relevant for understanding and controlling model behavior. 

\citet{sofroniew2026emotion} recently found that Claude Sonnet 4.5 contains
internal representations of emotional concepts that linearly predict
and causally influence behavior, which they term functional
emotions. This is an important result yet it leaves three questions open.
First, the evidence is primarily correlational. Probing and steering demonstrate that emotion-like features exist but do not
isolate which specific layers are causally responsible.
Second, the study focuses on a single closed frontier model.
Third, it does not test whether the observed signal reflects genuine
valence tracking or topic detection. A model that simply recognizes
the word ``rejected'' could produce the same probing signal as one
with a rich internal valence representation. Without causal localization, it remains unclear whether emotional representations are concentrated within specific processing stages or distributed diffusely across the network.
 
We address these gaps using activation patching~\citep{vig2020causal, meng2022locating} on three open-source LLMs together with a prompt design that explicitly controls for topic confounds. Our results show a clear depth-wise dissociation in valence processing: negative outcomes are causally localized to early layers while positive outcomes peak at mid-to-late layers. This pattern is consistent across all evaluated models. We further validate that the effect is genuinely valence-specific through a topic-controlled flip test. Using a shared-corrupted-baseline design, prompt pairs reverse sign significantly above chance when valence changes. Finally, causal steering experiments demonstrate that the identified representations behave as manipulable linear directions. Adding or subtracting them from the residual stream predictably alters responses to emotionally neutral inputs.

\section{Related Work}
\label{sec:related}

\paragraph{Mechanistic interpretability.}
Activation patching has been widely used to study factual recall and causal circuits in transformers~\citep{meng2022locating, wang2022interpretability, conmy2023towards}. 
\citet{lieberum2023does} examine whether circuit analysis techniques scale to larger models, highlighting challenges in extending interpretability methods beyond small settings. 
The residual stream framework~\citep{elhage2021mathematical} and logit lens~\citep{nostalgebraist2020logit} provide theoretical grounding for layer-wise analysis of internal representations.

\paragraph{Emotion and sentiment in LLMs.}
\citet{tigges2023linear} and \citet{marks2023geometry} identify linearly decodable sentiment and truth directions in intermediate transformer layers using probing methods. \citet{tak2025mechanistic} provide a mechanistic analysis of emotion inference across model families, finding concentration of emotion-relevant computations in specific layers. 
\citet{zhang2025decoding} further separate affect reception and emotion categorization across depth, suggesting functional specialization across layers. 
\citet{wang2025llms} study emotion circuits in LLMs by identifying neurons and attention heads associated with emotional expression and demonstrate controllable emotion generation through circuit-level interventions. \citet{hofmann2024dialect} show that transformer representations encode socially relevant attributes beyond surface text patterns.

\paragraph{Steering vectors.}
Steering vectors have been shown to correspond to directions in activation space that reliably control model behavior~\citep{turner2024activation, zou2023representation}. 
They have been applied to identify and modify specific behaviors such as refusal~\citep{arditi2024refusal} and to study robustness across inputs~\citep{panickssery2023steering}.

\paragraph{Representation geometry.}
The linear representation hypothesis posits that semantic concepts are encoded as linear directions in activation space~\citep{park2023linear, elhage2022toy}. 
Recent work supports the existence of monosemantic or highly interpretable features at scale~\citep{bricken2023monosemanticity, templeton2024scaling}, suggesting structured geometry in learned representations.

Our work connects these lines by studying whether valence exhibits layer-dependent causal structure and whether it can be expressed as a manipulable linear direction in residual stream space.

\section{Methods}
\label{sec:methods}

\subsection{Methodological Background}
Our analysis builds on two core interpretability techniques: activation patching and steering vectors. Activation patching~\citep{vig2020causal, meng2022locating} is a causal intervention method for identifying where in a network information relevant to a behavior is computed. Given a clean and a corrupted input that differ in a controlled way, we run the model on the corrupted input while replacing residual stream activations at a given layer with those from the clean run. If this restores the clean output behavior, that layer is considered causally important for the computation. We sweep all layers and measure effects using a logit gap metric defined in Section~\ref{sec:metric}.

Steering vectors~\citep{turner2024activation, zou2023representation} are linear directions in activation space that can be added to residual stream activations at inference time to systematically shift model behavior. They are constructed from differences between activations under contrasting conditions and correspond to directions in representation space associated with latent features. We extract steering vectors from the residual stream at the most causally relevant layer identified by patching.

\subsection{Models}
 
We study three instruct-tuned models that span two architectural families and two scales: Llama-3.2-1B-Instruct \citep{llama3_2_modelcard}, Qwen2.5-1.5B-Instruct, and Qwen2.5-3B-Instruct \citep{qwen2_5_technical_report}. The Llama model employs Multi-head attention \citep{cordonnier2020multi} while the Qwen models use Grouped-query attention \citep{ainslie2023gqa}. Llama-1B vs.\ Qwen-1.5B provides an approximate architecture comparison (MHA vs.\ GQA), while Qwen-1.5B vs.\ Qwen-3B provides a scale comparison within the same attention architecture. Since Llama-1B and Qwen-1.5B differ in both architecture and parameter count, the architecture comparison is only approximate.
 
MHA uses independent key-value projections per head. GQA shares key-value projections across groups of heads, changing how information is routed between layers. We study whether the layer dissociation finding holds across both variants. All models use a fixed system prompt: ``You are a concise assistant. Respond in one or two sentences.''
All experiments use \texttt{float16} precision on a single GPU.
 
\subsection{Prompt Pair Design}
\label{sec:prompts}
 
We construct 100 clean/corrupted prompt pairs per condition.
Each pair consists of a clean prompt and a corrupted prompt that
differ in exactly one dimension: emotional valence.
We use two conditions:
 
\begin{itemize}
  \item \textbf{Good news:} the clean prompt describes a positive
    outcome; the corrupted prompt describes the same situation
    neutrally with no emotional signal.
  \item \textbf{Negative control:} the clean prompt describes a
    negative outcome; the corrupted prompt is \emph{identical} to
    the good-news corrupted baseline.
\end{itemize}
 
The shared corrupted baseline is the core design decision.
It means both conditions are patched into the exact same starting point, so any difference in patch effect can only be attributed to the valence of the clean prompt and not to topic or phrasing. A model doing pure topic detection would produce similar patch effects from both clean runs since the topic is the same. A model with a genuine internal valence representation would produce sign-opposite effects since the valence is opposite. Example pair:

\begin{tcolorbox}[promptbox]
\textcolor{teal}{\textbf{Clean (good news):}} ``I just got accepted into my dream PhD program today.'' \\[3pt]

\textcolor{red!70!black}{\textbf{Clean (negative):}} ``I just got rejected from my dream PhD program today.'' \\[3pt]

\textcolor{gray!60!black}{\textbf{Corrupted (shared):}} ``I just received an email about my PhD program today.''
\end{tcolorbox}
 
Pairs are drawn from three broad domains: academia, career and
personal life. These domains were chosen to represent situations where emotional outcomes are common and clearly valenced while covering distinct
vocabularies. If the layer dissociation held only in academia prompts it could reflect domain-specific vocabulary and not valence processing. We verify it holds across all three domains in Appendix~\ref{app:domains}.
 
Pairs where the clean and corrupted prompts tokenize to different lengths are left-padded to equal length using the model's pad token (or EOS as fallback). This preserves right-alignment of content tokens so the final token position is always a real token.

\subsection{Valence Metric}
\label{sec:metric}
 
We measure a model's emotional lean by computing a logit gap between
two sets of anchor tokens at the next-token distribution.
 
\paragraph{Anchor tokens.}
Positive anchors include \textit{congratulations, congrats, happy, glad, wonderful, amazing, thrilled, proud, fantastic, excellent} while negative anchors include \textit{okay, noted, fine, ordinary, received, sorry}.
We include only tokens that tokenize as a single unit for a given
model since multi-token words have split probability mass and
produce unreliable logit comparisons.
The number of valid anchors varies by model and is reported in
each experiment.
 
\paragraph{Score definition.}
\begin{equation}
  \text{score} = \frac{1}{|\mathcal{P}|} \sum_{i \in \mathcal{P}}
  \log p_i -
  \frac{1}{|\mathcal{N}|} \sum_{j \in \mathcal{N}} \log p_j
  \label{eq:metric}
\end{equation}
 
A positive score indicates the model's next-token distribution leans
toward positive-valence tokens. A negative score indicates a lean
toward negative-valence tokens. The score gap between clean and
corrupted runs measures how strongly the emotional content of the
clean prompt shifts the model's output distribution.
 
\paragraph{Metric validity.}
The metric is a proxy: it measures next-token probability mass on a fixed anchor set but does not directly measure an internal emotional state.
The main risks are anchor sensitivity, where different token sets can produce different scores, and ceiling effects, where models that strongly favor a specific token can artificially inflate the gap.
We address anchor sensitivity by running the same experiment with
three alternative anchor sets and reporting Spearman rank correlation
between the resulting score gaps (Appendix~\ref{app:anchors}).
High correlation indicates the findings are not specific to our
chosen anchors.
 
\subsection{Residual Stream Patching}
\label{sec:patching}
 
For each prompt pair we run the following procedure:
 
\begin{enumerate}
  \item Run the model on the clean tokens, caching
    \texttt{hook\_resid\_pre} activations at every layer.
    We cache only residual stream pre-activations instead of all intermediate activations to reduce memory overhead.
  \item Run the model on the corrupted tokens.
  \item For each layer $l \in \{0, \ldots, L-1\}$, replace the
    corrupted residual stream at layer $l$ with the cached clean
    activations and measure the resulting valence score.
  \item Record the patch effect at each layer: the change in
  valence score from patching at layer $l$.
\end{enumerate}
 
We record two summary statistics per prompt:
\begin{itemize}
  \item \texttt{top\_layer}: the layer index with the maximum patch
    effect, showing where the valence signal is most causally
    concentrated.
  \item \texttt{max\_patch\_effect}: the magnitude of the peak patch
    effect, reflecting how strongly a single layer drives the output.
\end{itemize}
 
The distribution of \texttt{top\_layer} across all prompts is the
primary evidence for the layer dissociation finding.
 
\subsection{Valence Flip Test}
\label{sec:fliptest}
 
The flip test directly tests whether the model tracks valence as a
separable variable or simply recognizes topic.
 
For each model, we align good-news and negative-control pairs by
index (they share the same corrupted baseline) and compute the score
gap for each pair under both conditions.
A flip occurs when:
\begin{equation}
  \text{gap}_\text{good\_news} > 0 \quad \text{and} \quad
  \text{gap}_\text{negative\_control} < 0
  \label{eq:flip}
\end{equation}
for the same prompt index.
 
A model that recognizes only topic would produce correlated gaps
across conditions, yielding a flip rate near 50\%.
A model with a genuine valence representation would produce
anti-correlated gaps, yielding a flip rate well above 50\%.
We report flip rate per model and use it as the primary evidence
against the topic detection hypothesis.
 
\subsection{Causal Steering}
\label{sec:steering}
 
The steering experiment tests whether the layer identified by
patching encodes valence as a manipulable linear direction or
merely correlates with valence-charged inputs.
 
\paragraph{Direction extraction.}
For each model and condition, we identify the top causal layer as
the median \texttt{top\_layer} across all 100 prompts.
We then extract a valence direction by computing the mean activation
difference between clean and corrupted runs at that layer across 50
randomly sampled prompt pairs:
 
\begin{equation}
  \vec{v} = \frac{1}{N} \sum_{i=1}^{N}
  \left( h^l_{\text{clean},i} - h^l_{\text{corr},i} \right),
  \qquad
  \hat{v} = \frac{\vec{v}}{\|\vec{v}\|}
  \label{eq:direction}
\end{equation}
 
where $h^l_{\cdot,i}$ denotes the last-token residual stream
activation at layer $l$ for the $i$-th prompt.
Unit-normalizing ensures the direction is scale-invariant.
 
\paragraph{Steering neutral prompts.}
We construct 50 emotionally neutral prompts. These are factual
context-free sentences with no emotional content (e.g.\ ``I went to
the library and borrowed a book today.'').
For each neutral prompt we apply the intervention:
 
\begin{equation}
  h^l \leftarrow h^l + \alpha \cdot \hat{v}
  \label{eq:intervention}
\end{equation}
 
at the last token position at layer $l$, sweeping
$\alpha \in \{-20, -10, -5, 0, 5, 10, 20\}$.
We measure the resulting valence score after the intervention and
compute $\Delta = \text{score}_\text{steered} - \text{score}_\text{base}$.

\section{Results}
\label{sec:results}

\subsection{Models Track Valence Directionally}
 
\begin{figure}[h]
  \centering
  \includegraphics[width=\linewidth, trim=0 0 0 35pt, clip]{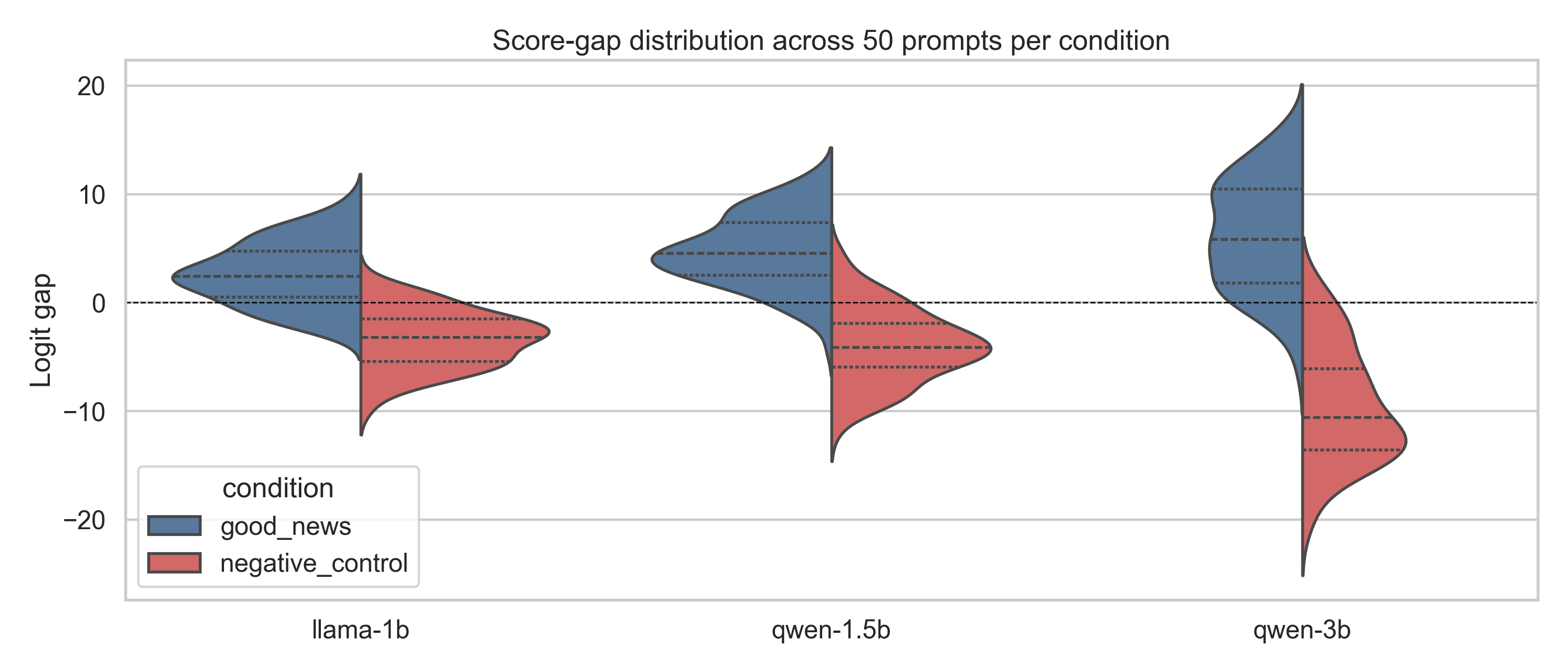}
  \caption{Score gap distributions across 100 prompts per condition
    for all three models. Each half-violin shows the distribution of
    logit gap scores: blue (left) for good-news prompts and red (right)
    for negative-control prompts. The two distributions are
    cleanly separated above and below zero in all three models.}
  \label{fig:violin}
\end{figure}

Before examining internal representations, we verify that all three
models actually produce directionally correct responses to emotional
content.
We compute the valence score (Eq.~\ref{eq:metric}) for each clean
and corrupted prompt pair and measure sign accuracy: the fraction of prompts where the score gap has the correct sign. This is a necessary baseline because if a model cannot distinguish good news from bad news in its output distribution, there is nothing to explain mechanistically.
Figure~\ref{fig:violin} shows the full score gap distributions
across all 100 prompts per condition which confirms the separation is
consistent and not driven by a few outliers.
 
Both model families respond correctly to valence across the majority
of prompts (Table~\ref{tab:scores}).
 
\begin{table}[h]
\centering
\small
\begin{tabular}{llcc}
\toprule
Model & Condition & Mean gap & Sign accuracy \\
\midrule
Llama-1B  & good\_news        & $+2.66$ & 79.0\% \\
Llama-1B  & negative\_control & $-3.39$ & 88.0\% \\
Qwen-1.5B & good\_news        & $+4.79$ & 92.0\% \\
Qwen-1.5B & negative\_control & $-3.97$ & 86.0\% \\
Qwen-3B   & good\_news        & $+6.27$ & 89.0\% \\
Qwen-3B   & negative\_control & $-9.45$ & 95.0\% \\
\bottomrule
\end{tabular}
\caption{Mean score gap (Eq.~\ref{eq:metric}) and sign accuracy
per model and condition. Positive gap = lean toward positive anchor
tokens; negative gap = lean toward negative anchors.}
\label{tab:scores}
\end{table}
 
Sign accuracy ranges from 79\% to 95\% across models and conditions.
The Qwen family produces sharper score gaps than Llama-1B, with
Qwen-3B showing the strongest negative-control signal (mean
gap $-9.45$, sign accuracy 95\%). Llama-1B's good-news sign accuracy of 79\% is the weakest result in the table which suggests positive valence tracking is noisier in the smaller model.

\subsection{The Flip Test Rules Out Topic Detection}
 
High sign accuracy is a necessary but not sufficient check.
A model that recognizes ``PhD program'' as emotionally loaded
could produce a positive gap for both conditions if topic alone
biases the output. The flip test directly targets this alternative by checking whether the score gap changes sign when valence changes while the topic and
corrupted baseline are held fixed.
 
Valence flip rates are: Llama-1B 69\%, Qwen-1.5B 80\% and Qwen-3B
85\%, all well above the 50\% chance baseline.
These rates are 19--35 percentage points above chance, making the margin substantial and unlikely to be attributable to noise.
Sign accuracy tells us the model responds correctly to valence while the flip test tells us the response is driven specifically by valence and not by topic, syntax or any other confound shared across conditions.
The consistency across Llama-1B (MHA) and Qwen models (GQA) further
suggests this is not an architecture-specific artifact.
Together, these results establish that the model encodes valence as
a separable internal variable, which is the precondition for the
layer dissociation analysis that follows.

\subsection{Negative and Positive Valence Localize at Different Depths}
 
Having confirmed that models respond directionally to valence, we
now ask where in the network this computation happens.
For each prompt pair we run residual stream patching across all
layers and record the layer with the peak patch effect
(\texttt{top\_layer}).
If they shared a single representation, top layers would cluster
at the same depth. A systematic difference indicates distinct
computational pathways.
 
\begin{figure}[h]
  \centering
  \includegraphics[width=\linewidth, trim=0 0 0 24pt, clip]{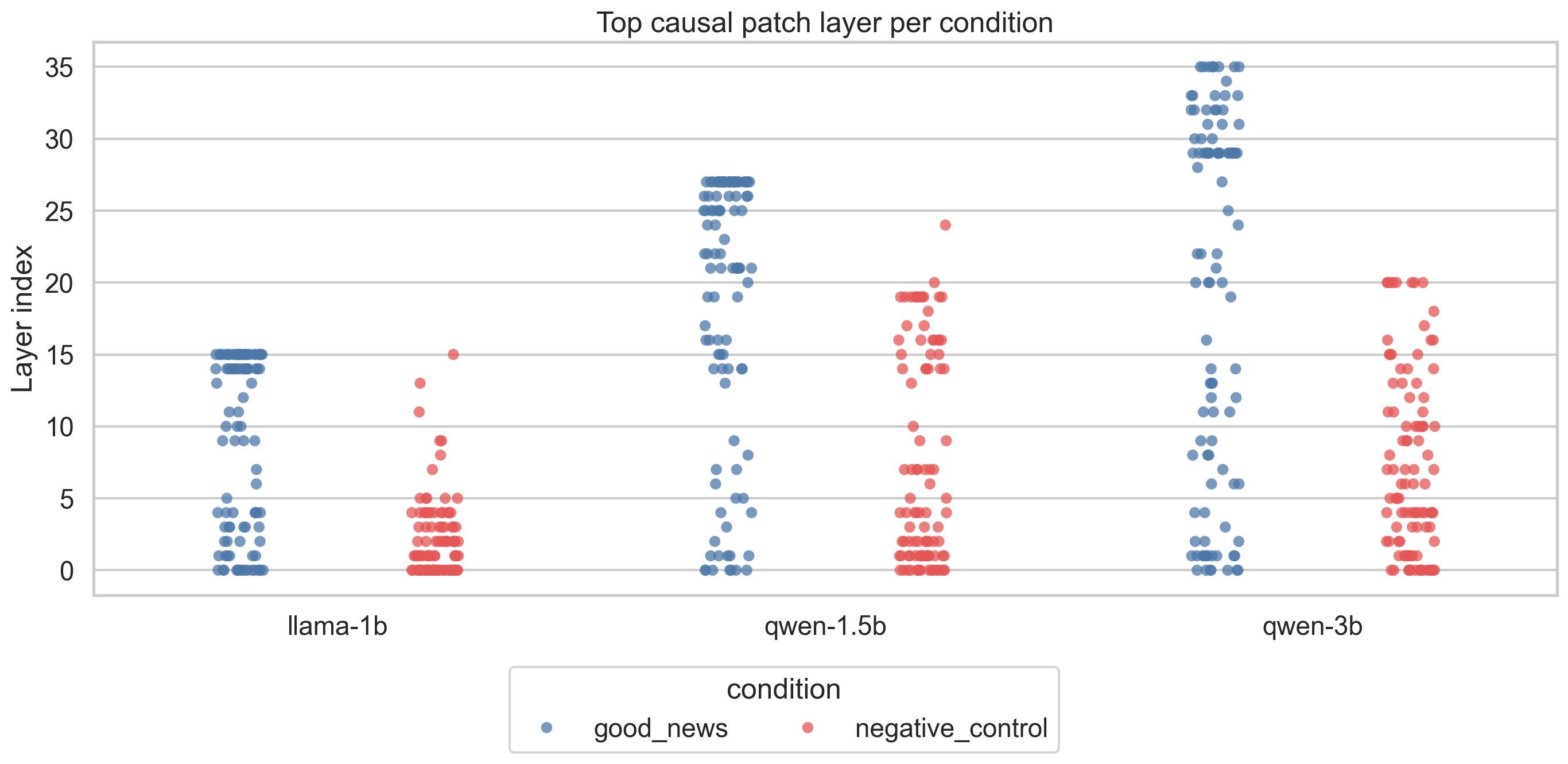}
  \caption{Top causal patch layer per prompt across all three models.
    Blue dots (good news) cluster at higher layers; red dots (negative
    control) cluster near the bottom. Each dot is one prompt; the
    dissociation is consistent and not driven by outliers.
    Good-news dots show wider vertical spread while negative-control dots pack tightly near the bottom, showing diffuse positive processing against sharply localized negative processing.}
  \label{fig:layers}
\end{figure}
 
This is the central finding of the paper.
Good-news signal peaks at 53--66\% of model depth; negative-control
signal peaks at 14--27\%.
Table~\ref{tab:mannwhitney} confirms this dissociation is highly
significant across all three models.
 
\begin{table}[h]
\centering
\small
\begin{tabular}{lccc}
\toprule
Model & $n_\text{layers}$ & $U$ & $p$-value \\
\midrule
Llama-1B  & 16 & 7808 & $1.84 \times 10^{-12}$ \\
Qwen-1.5B & 28 & 8071 & $2.58 \times 10^{-14}$ \\
Qwen-3B   & 36 & 7501 & $4.62 \times 10^{-10}$ \\
\bottomrule
\end{tabular}
\caption{Mann-Whitney $U$ test on \texttt{top\_layer} distributions:
good\_news vs.\ negative\_control. One-sided alternative: good\_news
top layers exceed negative\_control top layers. All three models are
highly significant which confirms the early-late dissociation is not
driven by chance or outliers.}
\label{tab:mannwhitney}
\end{table}
 
Early layers in transformers are generally associated with
low-level lexical processing~\citep{elhage2021mathematical}.
The fact that negative valence peaks there suggests the model has
a fast shallow detector for negative outcomes. Outcome words like
``rejected'' and ``failed'' are lexically distinctive and may not
require deep contextual integration to trigger a response.
Positive valence, by contrast, peaks at 53--66\% of model depth. This suggests that producing a warm response requires the model to first assemble a richer contextual representation of what the good outcome actually means.
This asymmetry is consistent across all three models regardless of
whether they use MHA or GQA attention. Domain-level analysis (Appendix~\ref{app:domains}) confirms the dissociation holds within academia, career and personal domains across all models ($p < 0.05$).

\subsection{Patch Magnitude Tracks Response Strength Asymmetrically}
 
Beyond identifying which layer peaks, we ask whether the size of
the patch effect predicts how strongly the model responds.
If valence is concentrated in one layer, a larger patch effect
should produce a larger output shift. We measure Spearman rank correlation between \texttt{max\_patch\_effect} and score gap across all prompts per model and condition. A strong correlation indicates localized processing while a weak one indicates the signal is distributed across many layers.
 
\begin{figure*}[t]
  \centering
  \includegraphics[width=\linewidth, trim= 0 0 0 30pt, clip]{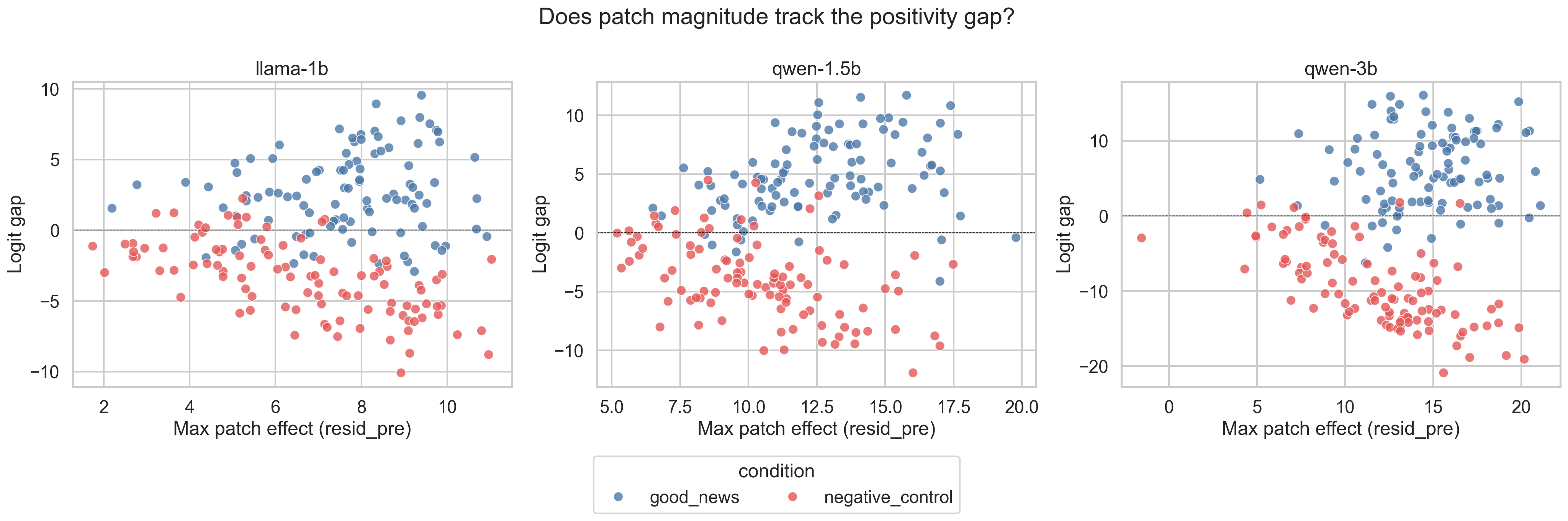}
  \caption{Max patch effect vs.\ score gap per prompt. For negative
    control (red), larger patch effects reliably predict more negative
    score gaps, with a single dominant layer driving the effect. For
    good news (blue) the relationship is weak, suggesting positive
    valence is more diffusely distributed across layers.}
  \label{fig:scatter}
\end{figure*}
 
Figure~\ref{fig:scatter} shows this asymmetry visually.
For negative control, patch magnitude strongly predicts response strength, with a single dominant layer driving the effect. The relationship collapses for good news indicative of positive valence being distributed across layers. Table~\ref{tab:spearman} reports the Spearman rank correlation
between patch magnitude and score gap per model and condition.
A large negative $\rho$ means that prompts where one layer dominates
the patch effect also tend to have more extreme score gaps, pointing
to tight causal localization. A $\rho$ near zero means the two are
unrelated, pointing to distributed processing.
 
\begin{table}[h]
\centering
\small
\begin{tabular}{llcc}
\toprule
Model & Condition & Spearman $\rho$ & $p$-value \\
\midrule
Llama-1B  & good\_news        & $+0.144$ & $0.15$ \\
Llama-1B  & negative\_control & $-0.646$ & $<10^{-12}$ \\
Qwen-1.5B & good\_news        & $+0.372$ & $1.37 \times 10^{-4}$ \\
Qwen-1.5B & negative\_control & $-0.488$ & $2.60 \times 10^{-7}$ \\
Qwen-3B   & good\_news        & $+0.112$ & $0.27$ \\
Qwen-3B   & negative\_control & $-0.676$ & $<10^{-13}$ \\
\bottomrule
\end{tabular}
\caption{Spearman $\rho$ between max patch effect and score gap.
Strong negative correlation for negative control points to a single
dominant layer driving the effect. Weak and non-significant
correlation for good news reflects diffuse processing across layers.}
\label{tab:spearman}
\end{table}
 
For negative control, patch magnitude is a reliable predictor of
response strength across all three models ($\rho$ between $-0.49$
and $-0.68$, all $p < 10^{-6}$). This means that when the model reads bad news, a single dominant layer drives the output. Its patch magnitude predicts how negative the response will be. However, for good news, this relationship is absent in Llama-1B and Qwen-3B ($\rho = 0.14$ and $0.11$) and only moderate in
Qwen-1.5B ($\rho = 0.37$). Positive valence does not route through a single dominant layer. The causal signal is spread across the network, which is why
no single patch location strongly predicts response strength.
 
\subsection{Steering Closes the Causal Loop}
Patching shows which layer carries the valence signal but does not
prove the residual stream at that layer encodes valence as a
manipulable direction. A layer could be causally active because it routes information instead of storing it. To test this, we extract a mean-difference valence direction at the top causal layer for each condition (Eq.~\ref{eq:direction}) and add it to the residual stream of 50
emotionally neutral prompts.
 
\begin{figure*}[t]
  \centering
  \includegraphics[width=\linewidth, trim=0 0 0 30pt, clip]{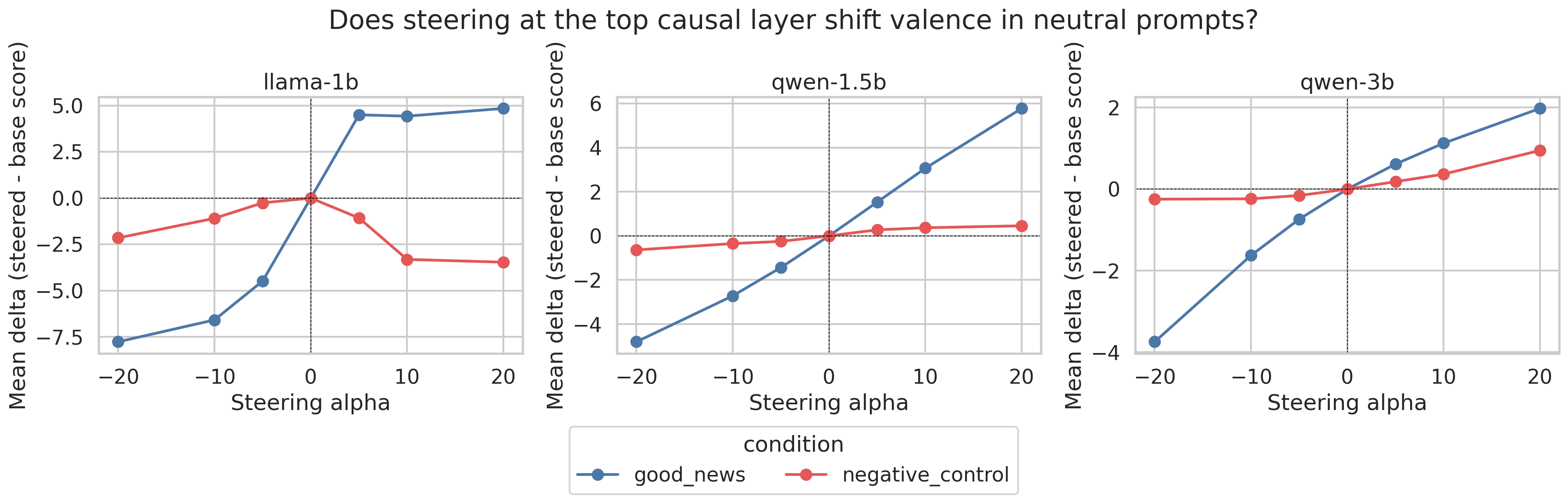}
  \caption{Mean change in valence score ($\Delta$ = steered minus base)
    as a function of steering strength $\alpha$, averaged over 50
    neutral prompts for each model.
    Blue line shows results for the good-news valence direction extracted
    at the good-news top layer. Red line shows results for the
    negative-control direction extracted at the negative-control top
    layer. The blue line rises monotonically confirming a clean
    steerable positive valence direction. The red line varies across
    models. Llama-1B shows a strongly inverted response while Qwen
    models show a weaker and flatter signal.}
  \label{fig:steering}
\end{figure*}
 
Steering at the patching-identified layers shifts neutral prompts
toward positive valence when the good-news direction is added across
all models (Table~\ref{tab:steering}).
The $\alpha$-$\Delta$ relationship is shown in Figure~\ref{fig:steering}.

\begin{table}[h]
\centering
\small
\begin{tabular}{lccc}
\toprule
Model & \% shifted (+) & \% shifted (-) & Mean $\Delta$ (+) \\
\midrule
Llama-1B  & 100\% & 70\% & $+4.43$ \\
Qwen-1.5B & 100\% & 82\% & $+3.07$ \\
Qwen-3B   & 92\%  & 84\% & $+1.13$ \\
\bottomrule
\end{tabular}
\caption{Steering results on 50 neutral prompts using the good-news
valence direction. (+) is the percentage of prompts that shift toward
positive valence at $\alpha=+10$. ($-$) is the percentage that shift
toward negative valence at $\alpha=-10$. Mean $\Delta$ is the average
change in valence score at $\alpha=+10$.}
\label{tab:steering}
\end{table}
 
At $\alpha = +10$, 100\% of Llama-1B prompts and 100\% of
Qwen-1.5B prompts shift positive, with Qwen-3B at 92\%.
The valence score increases by $+4.43$ for Llama-1B, $+3.07$ for
Qwen-1.5B and $+1.13$ for Qwen-3B at $\alpha=+10$ on prompts
with no emotional content.
At $\alpha = -10$, 70\%, 82\% and 84\% of prompts shift toward
negative valence for Llama-1B, Qwen-1.5B and Qwen-3B respectively. This 
shows that the good-news direction has genuine bidirectional control
over emotional lean.
The relationship between $\alpha$ and $\Delta$ is monotonically
increasing for the good-news direction in all three models
(Spearman $\rho > 0.89$, $p < 10^{-100}$). This confirms that the extracted direction is a genuine linear valence direction and not a noise vector.

\section{Discussion}
\label{sec:discussion}

\paragraph{The early-late asymmetry.}
The dissociation is the most surprising result.
It is loosely consistent with negativity bias in biological
systems~\citep{cacioppo1999affect, baumeister2001bad} where threats
are processed faster than rewards.
A simpler explanation is that negative outcome words (``rejected,''
``failed,'' ``denied'') are more lexically distinctive and detectable
from surface features alone, requiring less contextual integration
than positive outcomes.
Distinguishing these two explanations is an important direction for
future work.

\paragraph{Implications.}
The finding that negative and positive valence are processed at
different network depths has direct consequences for how we think
about emotion in LLMs.
It suggests these are not two ends of a single valence axis but
distinct computational processes that can in principle be targeted
independently.
The early negative-valence signal in particular is a natural
candidate for monitoring because it appears before the model has assembled
a full response and is predictably steerable.
 
\section{Limitations}
\label{sec:limitations}
 
Our metric is a proxy for valence sensitive to anchor choice.
Stability across three anchor sets (Appendix~\ref{app:anchors})
with $\rho$ between 0.47 and 0.87 indicates the findings are not
strongly anchor-driven, though the lower end of this range leaves
some residual uncertainty. The steering direction is extracted from the same prompt distribution used for patching and may overfit to it. Future work should validate on out-of-distribution emotional content. We study instruct-tuned models and whether the same asymmetry exists in base models remains an open question. The Llama-1B vs.\ Qwen-1.5B comparison is approximate because the models differ in both architecture and parameter count.

\section{Conclusion}
\label{sec:conclusion}
 
We have shown that emotional valence in LLMs is not a monolithic
internal state but an asymmetric computation. Negative valence
localizes causally to early layers while positive valence peaks at
mid-to-late layers.
A controlled flip test rules out topic detection as an alternative
explanation. A mean-difference valence direction extracted at those
layers shifts the vast majority of emotionally neutral prompts toward positive valence,
confirming the residual stream there encodes
valence as a genuinely manipulable representation.
Understanding why the two polarities require such different depths
of computation may reveal something fundamental about how
transformers encode semantic and emotional content.

\section*{Acknowledgements}
The authors thank RunPod for GPU compute resources used to run all
experiments in this work and Hugging Face for hosting the open-source
models used throughout.

\section*{Impact Statement}
This work studies the internal representations of emotional valence
in open-source language models.
We do not foresee direct negative societal impacts from this
research.
The findings may inform future work on monitoring internal model
states, which could contribute to safer and more interpretable
language model deployments.
All models studied are publicly available and the experiments do not
involve human subjects or sensitive data.

\section*{Disclosure of AI Usage}
Large language models were used for language editing of the draft
(Gemini) and code assistance (Claude Code) in this work.
All research ideas, technical content, experimental design, results
and conclusions are the original work of the authors.

\bibliography{example_paper}

@article{sofroniew2026emotion,
  title={Emotion concepts and their function in a large language model},
  author={Sofroniew, Nicholas and Kauvar, Isaac and Saunders, William and Chen, Runjin and Henighan, Tom and Hydrie, Sasha and Citro, Craig and Pearce, Adam and Tarng, Julius and Gurnee, Wes and others},
  journal={arXiv preprint arXiv:2604.07729},
  year={2026}
}

@inproceedings{tak2025mechanistic,
  title={Mechanistic interpretability of emotion inference in large language models},
  author={Tak, Ala N and Banayeeanzade, Amin and Bolourani, Anahita and Kian, Mina and Jia, Robin and Gratch, Jonathan},
  booktitle={Findings of the Association for Computational Linguistics: ACL 2025},
  pages={13090--13120},
  year={2025}
}

@article{wang2025llms,
  title={Do LLMs" Feel"? Emotion Circuits Discovery and Control},
  author={Wang, Chenxi and Zhang, Yixuan and Yu, Ruiji and Zheng, Yufei and Gao, Lang and Song, Zirui and Xu, Zixiang and Xia, Gus and Zhang, Huishuai and Zhao, Dongyan and others},
  journal={arXiv preprint arXiv:2510.11328},
  year={2025}
}

@article{zhang2025decoding,
  title={Decoding Emotion in the Deep: A Systematic Study of How LLMs Represent, Retain, and Express Emotion},
  author={Zhang, Jingxiang and Zhong, Lujia},
  journal={arXiv preprint arXiv:2510.04064},
  year={2025}
}

@article{meng2022locating,
  title={Locating and editing factual associations in gpt},
  author={Meng, Kevin and Bau, David and Andonian, Alex and Belinkov, Yonatan},
  journal={Advances in neural information processing systems},
  volume={35},
  pages={17359--17372},
  year={2022}
}

@article{vig2020causal,
  title={Causal mediation analysis for interpreting neural nlp: The case of gender bias},
  author={Vig, Jesse and Gehrmann, Sebastian and Belinkov, Yonatan and Qian, Sharon and Nevo, Daniel and Sakenis, Simas and Huang, Jason and Singer, Yaron and Shieber, Stuart},
  journal={arXiv preprint arXiv:2004.12265},
  year={2020}
}

@article{turner2024activation,
  title={Activation addition: Steering language models without optimization},
  author={Turner, Alexander Matt and Thiergart, Lisa and Leech, Gavin and Udell, David and Mini, Ulisse and MacDiarmid, Monte},
  year={2024}
}

@article{zou2023representation,
  title={Representation engineering: A top-down approach to ai transparency},
  author={Zou, Andy and Phan, Long and Chen, Sarah and Campbell, James and Guo, Phillip and Ren, Richard and Pan, Alexander and Yin, Xuwang and Mazeika, Mantas and Dombrowski, Ann-Kathrin and others},
  journal={arXiv preprint arXiv:2310.01405},
  year={2023}
}

@article{arditi2024refusal,
  title={Refusal in language models is mediated by a single direction},
  author={Arditi, Andy and Obeso, Oscar and Syed, Aaquib and Paleka, Daniel and Panickssery, Nina and Gurnee, Wes and Nanda, Neel},
  journal={Advances in Neural Information Processing Systems},
  volume={37},
  pages={136037--136083},
  year={2024}
}

@article{elhage2021mathematical,
  title={A mathematical framework for transformer circuits},
  author={Elhage, Nelson and Nanda, Neel and Olsson, Catherine and Henighan, Tom and Joseph, Nicholas and Mann, Ben and Askell, Amanda and Bai, Yuntao and Chen, Anna and Conerly, Tom and others},
  journal={Transformer Circuits Thread},
  volume={1},
  number={1},
  pages={12},
  year={2021}
}

@article{wang2022interpretability,
  title={Interpretability in the wild: a circuit for indirect object identification in gpt-2 small},
  author={Wang, Kevin and Variengien, Alexandre and Conmy, Arthur and Shlegeris, Buck and Steinhardt, Jacob},
  journal={arXiv preprint arXiv:2211.00593},
  year={2022}
}

@misc{nostalgebraist2020logit,
  title  = {Interpreting {GPT}: the Logit Lens},
  author = {nostalgebraist},
  year   = {2020},
  url    = {https://www.lesswrong.com/posts/AcKRB8wDpdaN6v6ru/interpreting-gpt-the-logit-lens}
}

@article{elhage2022toy,
  title={Toy models of superposition},
  author={Elhage, Nelson and Hume, Tristan and Olsson, Catherine and Schiefer, Nicholas and Henighan, Tom and Kravec, Shauna and Hatfield-Dodds, Zac and Lasenby, Robert and Drain, Dawn and Chen, Carol and others},
  journal={arXiv preprint arXiv:2209.10652},
  year={2022}
}

@article{templeton2024scaling,
       title={Scaling Monosemanticity: Extracting Interpretable Features from Claude 3 Sonnet},
       author={Templeton, Adly and Conerly, Tom and Marcus, Jonathan and Lindsey, Jack and Bricken, Trenton and Chen, Brian and Pearce, Adam and Citro, Craig and Ameisen, Emmanuel and Jones, Andy and Cunningham, Hoagy and Turner, Nicholas L and McDougall, Callum and MacDiarmid, Monte and Freeman, C. Daniel and Sumers, Theodore R. and Rees, Edward and Batson, Joshua and Jermyn, Adam and Carter, Shan and Olah, Chris and Henighan, Tom},
       year={2024},
       journal={Transformer Circuits Thread},
       url={https://transformer-circuits.pub/2024/scaling-monosemanticity/index.html}
    }

@article{park2023linear,
  title={The linear representation hypothesis and the geometry of large language models},
  author={Park, Kiho and Choe, Yo Joong and Veitch, Victor},
  journal={arXiv preprint arXiv:2311.03658},
  year={2023}
}

@article{tigges2023linear,
  title={Linear representations of sentiment in large language models},
  author={Tigges, Curt and Hollinsworth, Oskar John and Geiger, Atticus and Nanda, Neel},
  journal={arXiv preprint arXiv:2310.15154},
  year={2023}
}

@article{marks2023geometry,
  title={The geometry of truth: Emergent linear structure in large language model representations of true/false datasets},
  author={Marks, Samuel and Tegmark, Max},
  journal={arXiv preprint arXiv:2310.06824},
  year={2023}
}

@article{panickssery2023steering,
  title={Steering llama 2 via contrastive activation addition},
  author={Panickssery, Nina and Gabrieli, Nick and Schulz, Julian and Tong, Meg and Hubinger, Evan and Turner, Alexander Matt},
  journal={arXiv preprint arXiv:2312.06681},
  year={2023}
}

@article{hofmann2024dialect,
  title={Dialect prejudice predicts AI decisions about people's character, employability, and criminality},
  author={Hofmann, Valentin and Kalluri, Pratyusha Ria and Jurafsky, Dan and King, Sharese},
  journal={arXiv preprint arXiv:2403.00742},
  year={2024}
}

@article{cacioppo1999affect,
  title={The affect system has parallel and integrative processing components: Form follows function.},
  author={Cacioppo, John T and Gardner, Wendi L and Berntson, Gary G},
  journal={Journal of personality and Social Psychology},
  volume={76},
  number={5},
  pages={839},
  year={1999},
  publisher={American Psychological Association}
}

@article{baumeister2001bad,
  title={Bad is stronger than good},
  author={Baumeister, Roy F and Bratslavsky, Ellen and Finkenauer, Catrin and Vohs, Kathleen D},
  journal={Review of general psychology},
  volume={5},
  number={4},
  pages={323--370},
  year={2001},
  publisher={SAGE Publications Sage CA: Los Angeles, CA}
}

@article{conmy2023towards,
  title={Towards automated circuit discovery for mechanistic interpretability},
  author={Conmy, Arthur and Mavor-Parker, Augustine and Lynch, Aengus and Heimersheim, Stefan and Garriga-Alonso, Adri{\`a}},
  journal={Advances in Neural Information Processing Systems},
  volume={36},
  pages={16318--16352},
  year={2023}
}

@article{olsson2022context,
  title={In-context learning and induction heads},
  author={Olsson, Catherine and Elhage, Nelson and Nanda, Neel and Joseph, Nicholas and DasSarma, Nova and Henighan, Tom and Mann, Ben and Askell, Amanda and Bai, Yuntao and Chen, Anna and others},
  journal={arXiv preprint arXiv:2209.11895},
  year={2022}
}

@article{lieberum2023does,
  title={Does circuit analysis interpretability scale? evidence from multiple choice capabilities in chinchilla},
  author={Lieberum, Tom and Rahtz, Matthew and Kram{\'a}r, J{\'a}nos and Nanda, Neel and Irving, Geoffrey and Shah, Rohin and Mikulik, Vladimir},
  journal={arXiv preprint arXiv:2307.09458},
  year={2023}
}

@article{bricken2023monosemanticity,
       title={Towards Monosemanticity: Decomposing Language Models With Dictionary Learning},
       author={Bricken, Trenton and Templeton, Adly and Batson, Joshua and Chen, Brian and Jermyn, Adam and Conerly, Tom and Turner, Nick and Anil, Cem and Denison, Carson and Askell, Amanda and Lasenby, Robert and Wu, Yifan and Kravec, Shauna and Schiefer, Nicholas and Maxwell, Tim and Joseph, Nicholas and Hatfield-Dodds, Zac and Tamkin, Alex and Nguyen, Karina and McLean, Brayden and Burke, Josiah E and Hume, Tristan and Carter, Shan and Henighan, Tom and Olah, Christopher},
       year={2023},
       journal={Transformer Circuits Thread},
       note={https://transformer-circuits.pub/2023/monosemantic-features/index.html}
    }

@misc{llama3_2_modelcard,
  title        = {Llama 3.2: Model Cards and Prompt Formats},
  author       = {Meta},
  year         = {2024},
  howpublished = {\url{https://www.llama.com/docs/model-cards-and-prompt-formats/llama3_2/}},
}

@article{qwen2_5_technical_report,
  title   = {Qwen2.5 Technical Report},
  author  = {Yang, An and Yang, Baosong and Zhang, Beichen and Hui, Binyuan and Zheng, Bo and Yu, Bowen and Li, Chengyuan and Liu, Dayiheng and Huang, Fei and Wei, Haoran and others},
  journal = {arXiv preprint arXiv:2412.15115},
  year    = {2024}
}

@inproceedings{ainslie2023gqa,
  title={Gqa: Training generalized multi-query transformer models from multi-head checkpoints},
  author={Ainslie, Joshua and Lee-Thorp, James and De Jong, Michiel and Zemlyanskiy, Yury and Lebr{\'o}n, Federico and Sanghai, Sumit},
  booktitle={Proceedings of the 2023 Conference on Empirical Methods in Natural Language Processing},
  pages={4895--4901},
  year={2023}
}

@article{cordonnier2020multi,
  title={Multi-head attention: Collaborate instead of concatenate},
  author={Cordonnier, Jean-Baptiste and Loukas, Andreas and Jaggi, Martin},
  journal={arXiv preprint arXiv:2006.16362},
  year={2020}
}
\bibliographystyle{icml2026}

\newpage
\appendix
\onecolumn

\section{Anchor Sensitivity}
\label{app:anchors}
 
We verify the valence metric (Eq.~\ref{eq:metric}) is stable across
three anchor sets.
The metric measures emotional lean by computing a logit gap between
a fixed set of positive and negative response tokens.
A concern with any such metric is that the findings could be specific to the particular tokens chosen. Different tokens might rank prompts differently, changing which prompts appear to have strong valence signal and which do not.
To test this, we define two alternative anchor sets (Alt1, Alt2) that
cover the same emotional territory using lexically distinct tokens
and repeat the score gap computation for each.
If the findings are genuine, score gaps computed with different anchor
sets should rank prompts similarly, producing high Spearman $\rho$.
 
Table~\ref{tab:anchor_unified} reports both the mean score gap and
Spearman $\rho$ between each pair of anchor sets.
Several patterns are worth noting.
First, all three sets produce sign-consistent mean gaps throughout. Every model under good-news shows a positive mean gap and every model under negative-control shows a negative mean gap, regardless of which anchor set is used. The sign of the effect is therefore not anchor-dependent.
Second, rank correlations between sets are moderate to strong
($\rho$ between 0.47 and 0.87), with most values exceeding 0.6.
The strongest correlations are between Alt1 and Alt2, suggesting
these two sets are more similar to each other than either is to the
default set. Third, the lowest correlations occur for Llama-1B good-news
(default vs.\ Alt1: $\rho = 0.47$).
This reflects the weaker overall good-news signal in Llama-1B, where noisier signals produce less stable rank orderings across anchor sets. Notably, Llama-1B's negative-control condition is more stable
($\rho$ between 0.50 and 0.79), consistent with the stronger and
more localized negative valence signal observed in the main results.
 
\begin{table}[h]
\centering
\small
\setlength{\tabcolsep}{4pt}
\begin{tabular}{llcccccc}
\toprule
 & & \multicolumn{3}{c}{Mean score gap} & \multicolumn{3}{c}{Spearman $\rho$} \\
\cmidrule(lr){3-5}\cmidrule(lr){6-8}
Model & Condition & Default & Alt1 & Alt2 & D vs A1 & D vs A2 & A1 vs A2 \\
\midrule
Llama-1B  & good\_news        & $+2.74$ & $+2.41$ & $+3.01$ & 0.647 & 0.744 & 0.867 \\
Llama-1B  & negative\_control & $-3.49$ & $-5.15$ & $-5.63$ & 0.503 & 0.693 & 0.791 \\
Qwen-1.5B & good\_news        & $+4.82$ & $+3.11$ & $+4.98$ & 0.472 & 0.690 & 0.768 \\
Qwen-1.5B & negative\_control & $-3.96$ & $-9.15$ & $-7.68$ & 0.631 & 0.585 & 0.840 \\
Qwen-3B   & good\_news        & $+6.95$ & $+6.25$ & $+8.06$ & 0.673 & 0.709 & 0.873 \\
Qwen-3B   & negative\_control & $-9.16$ & $-20.19$ & $-15.20$ & 0.746 & 0.622 & 0.806 \\
\bottomrule
\end{tabular}
\caption{Anchor sensitivity table showing mean score gap and
Spearman $\rho$ between all pairs of anchor sets per model and
condition. Mean score gap columns confirm the sign of the effect is
the same across all three anchor sets: positive for good news and
negative for negative control throughout. Spearman $\rho$ columns
report rank correlations between sets ($\rho$ between 0.47 and 0.87) which
shows that prompt orderings are broadly stable across anchor
choices.}
\label{tab:anchor_unified}
\end{table}
 
\noindent The anchor sets used are:
\begin{itemize}
  \item \textbf{Default:} congratulations, happy, glad, wonderful, amazing, thrilled, proud, fantastic, excellent $|$ okay, noted, fine, ordinary, received, sorry
  \item \textbf{Alt1:} great, awesome, excited, delighted, superb, impressive, brilliant, pleased, overjoyed, lucky $|$ unfortunately, sadly, regret, disappointing, difficult, tough
  \item \textbf{Alt2:} celebrate, congratulate, achievement, success, earned, deserve, outstanding, remarkable, triumph $|$ failed, rejected, denied, missed, lost, unfortunate
\end{itemize}

\section{Domain Analysis}
\label{app:domains}
 
The domain analysis tests whether the layer dissociation finding
holds within each domain independently.
If the finding were driven by topic-specific vocabulary, for example
if academic words like ``accepted'' and ``rejected'' happened to
produce early-layer responses due to their frequency in training
data, we would expect the finding to disappear within career or
personal prompts where the vocabulary is entirely different.
Conversely, if the dissociation holds across all three domains, it
provides strong evidence that the model is responding to valence as
a general property and not to any single domain's lexical patterns.
 
\paragraph{Representative prompt pairs.}
Table~\ref{tab:represent} shows one clean/corrupted pair from each domain.
The corrupted prompt is emotionally neutral and identical across
good-news and negative-control conditions within each domain.
This design ensures that any difference in patch effect between conditions within a domain is attributable to valence alone and not to phrasing or topic differences.
The prompts were written to be realistic and emotionally clear while
avoiding artificially extreme language.
 
\begin{table}[h]
\centering
\small
\begin{tabular}{p{1.8cm}p{5.5cm}p{3.5cm}}
\toprule
Domain & Clean prompt & Corrupted prompt \\
\midrule
Academia &
  \textcolor{teal}{\textit{Good:} ``I just got accepted into my dream PhD program today.''} \newline
  \textcolor{red!70!black}{\textit{Negative:} ``I just got rejected from my dream PhD program today.''} &
  \textcolor{gray!60!black}{``I just received an email about my PhD program today.''} \\
\addlinespace
Career &
  \textcolor{teal}{\textit{Good:} ``My contract was renewed with a significant pay increase.''} \newline
  \textcolor{red!70!black}{\textit{Negative:} ``My contract was not renewed and I was let go this week.''} &
  \textcolor{gray!60!black}{``My contract renewal came through this week.''} \\
\addlinespace
Personal &
  \textcolor{teal}{\textit{Good:} ``My biopsy results came back completely benign.''} \newline
  \textcolor{red!70!black}{\textit{Negative:} ``My biopsy results came back positive and I need further treatment.''} &
  \textcolor{gray!60!black}{``I received my biopsy results from the clinic.''} \\
\bottomrule
\end{tabular}
\caption{Representative prompt pairs from each domain.
The corrupted prompt is emotionally neutral and shared across both
conditions. Only the clean prompt varies in valence.}
\label{tab:represent}
\end{table}
 
\begin{figure}[h]
  \centering
  \includegraphics[width=\linewidth, trim=0 0 0 50pt, clip]{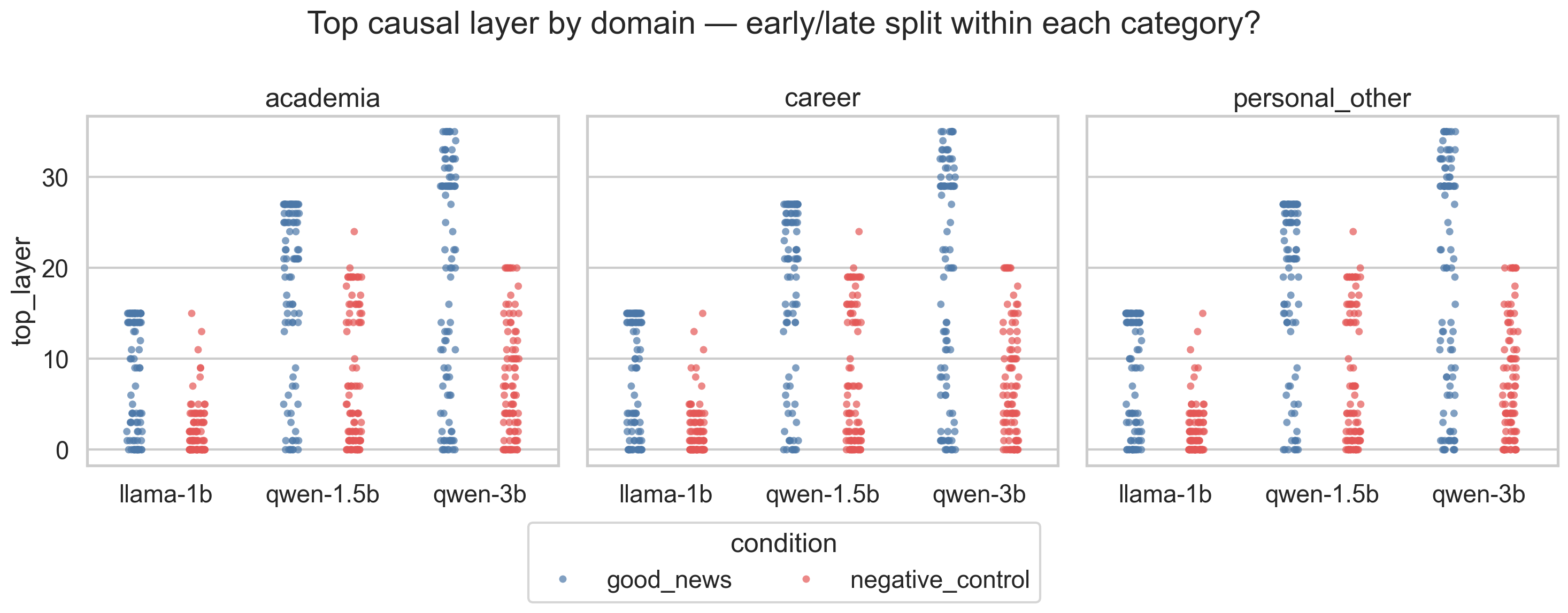}
  \caption{Top causal patch layer per prompt, broken down by domain.
    Blue dots (good news) cluster higher on the y-axis than red dots
    (negative control) within academia, career and personal domains
    independently. Each panel replicates the early-late dissociation
    seen in the main results within a single domain, ruling out the
    possibility that the finding is driven by vocabulary specific to
    any one topic area. The vertical spread within each condition
    reflects natural variation in which layer peaks for individual
    prompts.}
  \label{fig:domain_layer}
\end{figure}

\paragraph{Results.}
Table~\ref{tab:domain} reports Mann-Whitney $U$ $p$-values for the layer
dissociation within each domain.
Academia and personal are highly significant across all three models
($p < 10^{-4}$), confirming the early-late dissociation is not
confined to a single topic area.
Career is significant for Qwen-1.5B ($p = 1.7 \times 10^{-3}$)
and borderline for Llama-1B and Qwen-3B.
The weaker career result is not surprising. Career outcome prompts tend to use more formal and hedged language than academia or personal prompts, which may reduce the clarity of the valence signal.
The key result is that at least two of the three domains are
independently significant in every model, ruling out the possibility
that the main finding is driven by a single domain.
 
\begin{table}[h]
\centering
\small
\begin{tabular}{lccc}
\toprule
Domain & Llama-1B & Qwen-1.5B & Qwen-3B \\
\midrule
academia      & $7.9\times10^{-6}$ & $6.3\times10^{-5}$ & $9.7\times10^{-5}$ \\
career        & $3.2\times10^{-2}$ & $1.7\times10^{-3}$ & $6.9\times10^{-2}$ \\
personal      & $1.9\times10^{-7}$ & $3.7\times10^{-8}$ & $6.5\times10^{-8}$ \\
\bottomrule
\end{tabular}
\caption{Mann-Whitney $U$ $p$-values testing whether good-news
top layers are significantly higher than negative-control top layers
within each domain. Academia and personal are highly significant
across all three models ($p < 10^{-4}$), confirming the early-late
dissociation holds within each domain independently. Career is
significant for Qwen-1.5B and borderline for Llama-1B and Qwen-3B,
likely due to more hedged language in career prompts reducing the
clarity of the valence signal.}
\label{tab:domain}
\end{table}

\section{Patch Effect Visualizations}
\label{app:qualitative}
 
Residual stream patch heatmaps provide a per-prompt view of where
the causal signal lives across layers and token positions.
Each cell in the heatmap corresponds to one layer (y-axis) and one
token position (x-axis).
The value in each cell is the patch effect at that layer and
position: how much the valence score changes when the clean
activation at that layer and position is patched into the corrupted
run.
Brighter (yellow/green) cells indicate larger positive patch effects.
Darker (blue/purple) cells indicate smaller or negative effects.
 
The key comparison is between the good-news and negative-control
heatmaps for the same model.
The layer dissociation predicts that good-news signal should
concentrate at mid-to-late layers while negative-control signal
should concentrate near the bottom of the network. The token position axis also provides interpretable structure, with signal tending to concentrate around the outcome phrase instead of around neutral context tokens.
We show one personal-domain prompt pair per model below.

\begin{tcolorbox}[promptbox, title=Qwen-1.5B Inputs (Personal)]
  \textcolor{teal}{\textbf{Good news:}} ``I passed my professional certification exam on the first attempt.'' \\[3pt]
  \textcolor{red!70!black}{\textbf{Negative control:}} ``I failed my language proficiency certification after months of preparation.'' \\[3pt]
  \textcolor{gray!60!black}{\textbf{Corrupted:}} ``I completed my professional certification exam today.''
\end{tcolorbox}
 
\begin{figure}[h]
\centering
\begin{minipage}{0.48\linewidth}
  \centering
  \includegraphics[width=\linewidth, trim=0 0 0 36.3pt, clip]{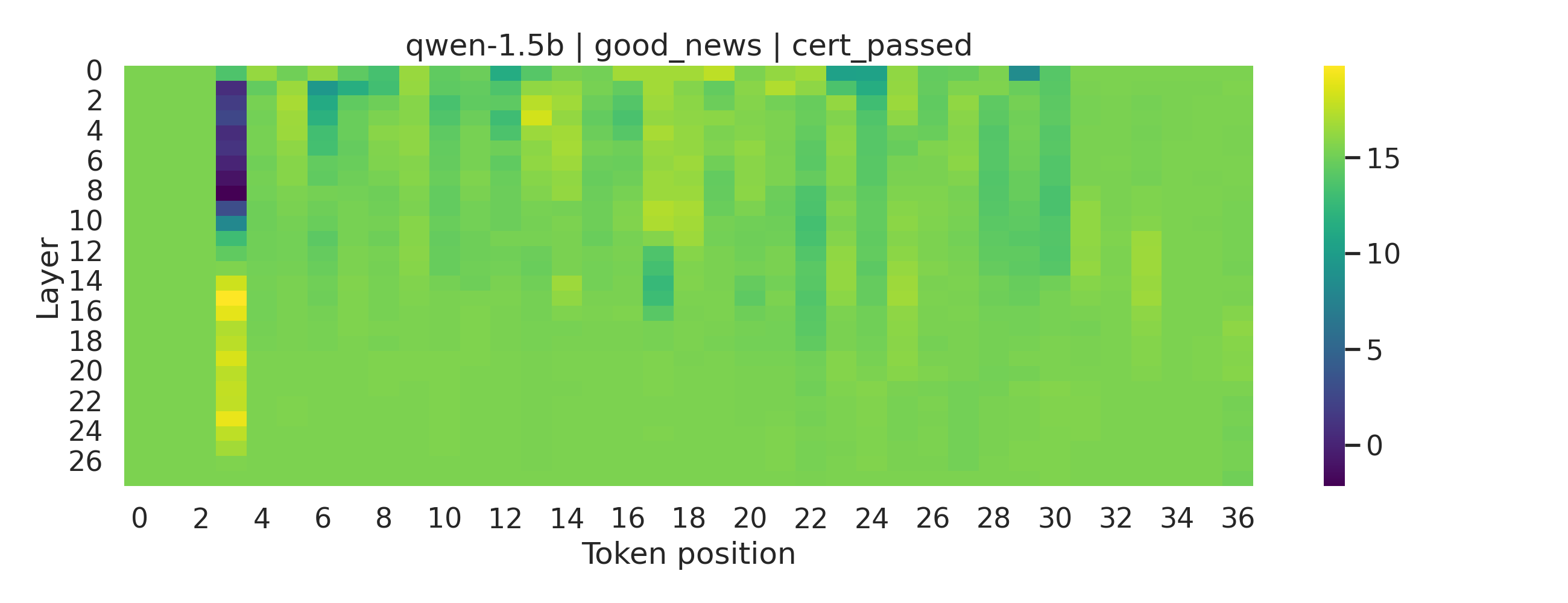}
\end{minipage}
\hfill
\begin{minipage}{0.48\linewidth}
  \centering
  \includegraphics[width=\linewidth, trim=0 0 0 36.3pt, clip]{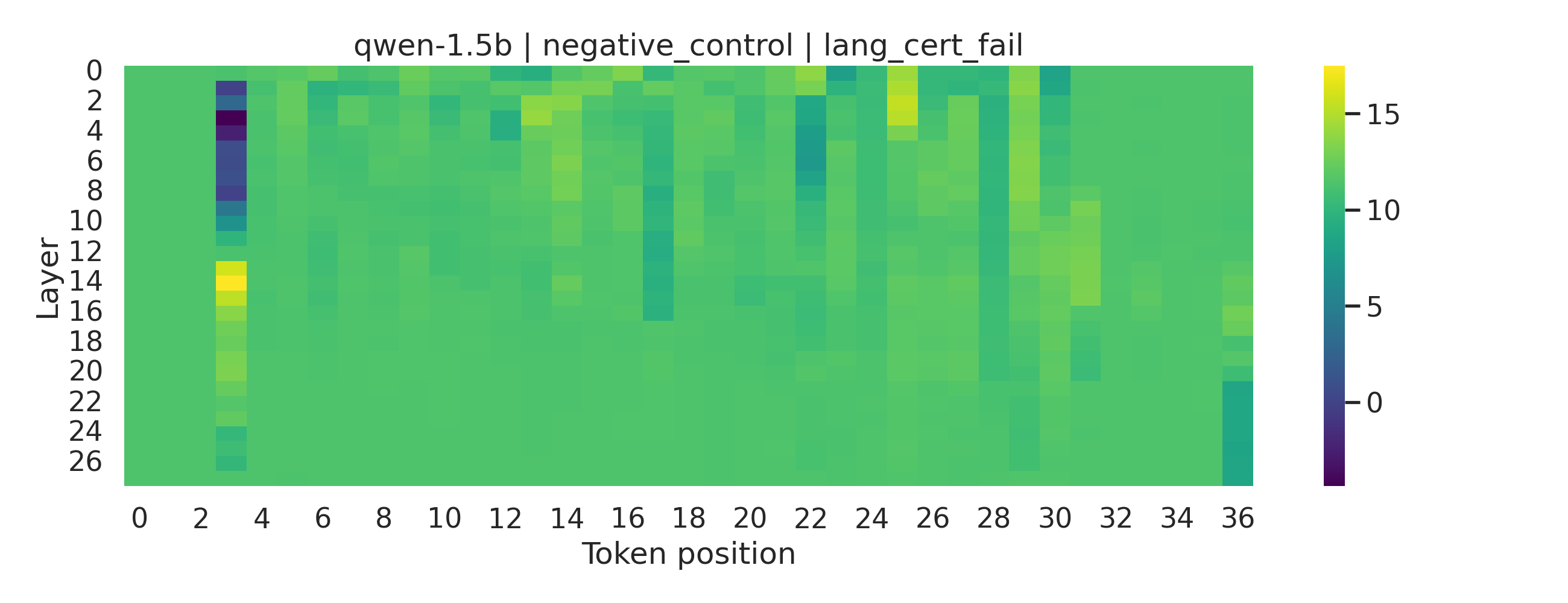}
\end{minipage}
\caption{Qwen-1.5B residual stream patch heatmaps. Good news (left)
  shows signal spread across mid-to-late layers. Negative control
  (right) shows activity concentrated in layers 0--5, consistent
  with the layer dissociation pattern observed in Qwen-3B.}
\label{fig:heatmaps_qwen15}
\end{figure}

\begin{tcolorbox}[promptbox, title=Qwen-3B Inputs (Personal)]
  \textcolor{teal}{\textbf{Good news:}} ``I passed my professional certification exam on the first attempt.'' \\[3pt]
  \textcolor{red!70!black}{\textbf{Negative control:}} ``I was rejected from my first-choice university with no feedback.'' \\[3pt]
  \textcolor{gray!60!black}{\textbf{Corrupted:}} ``I completed my professional certification exam today.''
\end{tcolorbox}
 
\begin{figure}[h]
\centering
\begin{minipage}{0.48\linewidth}
  \centering
  \includegraphics[width=\linewidth, trim=0 0 0 36.5pt, clip]{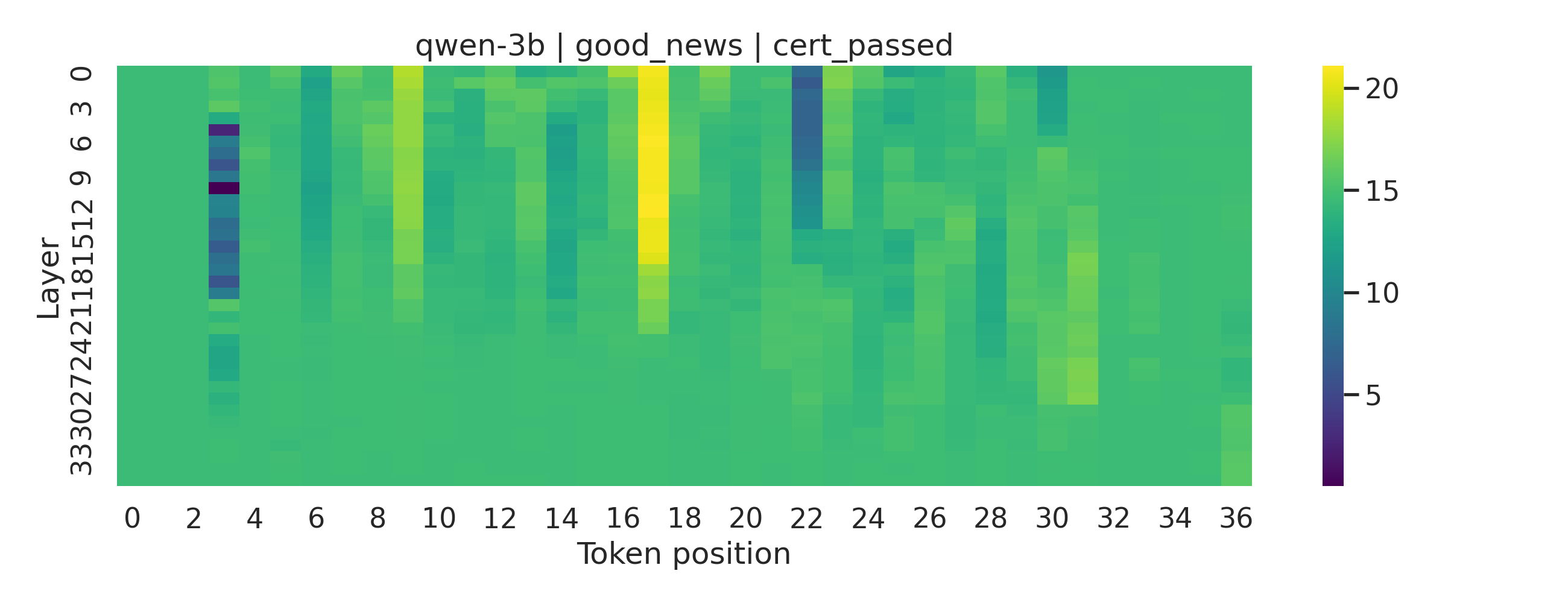}
\end{minipage}
\hfill
\begin{minipage}{0.48\linewidth}
  \centering
  \includegraphics[width=\linewidth, trim=0 0 0 36.5pt, clip]{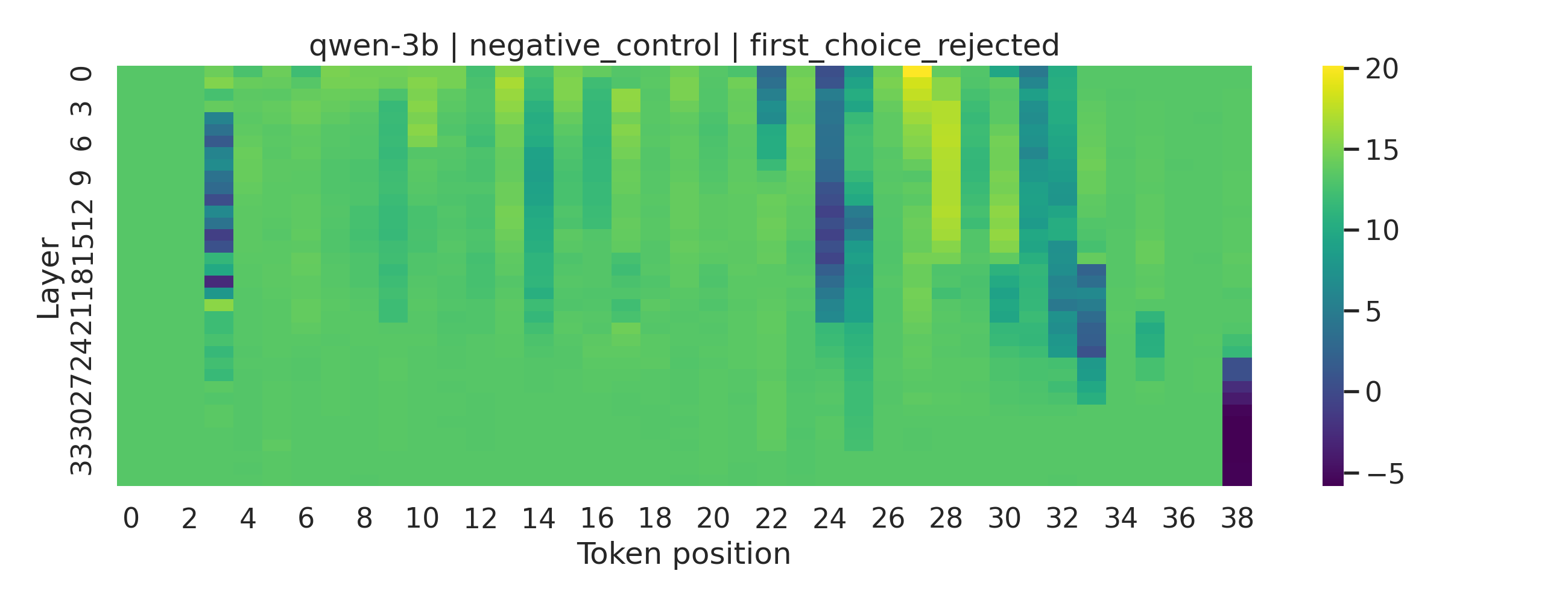}
\end{minipage}
\caption{Qwen-3B residual stream patch heatmaps. Good news (left)
  shows bright cells in the upper half of the y-axis, concentrated
  around the outcome phrase tokens. Negative control (right) shows
  stronger activity in the bottom rows, directly visualizing the
  layer dissociation.}
\label{fig:heatmaps_qwen}
\end{figure}
 
\begin{tcolorbox}[promptbox, title=Llama-1B Inputs (Personal)]
  \textcolor{teal}{\textbf{Good news:}} ``I completed the hardest hiking trail in my region without stopping.'' \\[3pt]
  \textcolor{red!70!black}{\textbf{Negative control:}} ``I gained back all the weight I had lost over six months.'' \\[3pt]
  \textcolor{gray!60!black}{\textbf{Corrupted:}} ``I completed a long hiking trail in my region today.''
\end{tcolorbox}
 
\begin{figure}[h]
\centering
\begin{minipage}{0.48\linewidth}
  \centering
  \includegraphics[width=\linewidth, trim=0 0 0 36.5pt, clip]{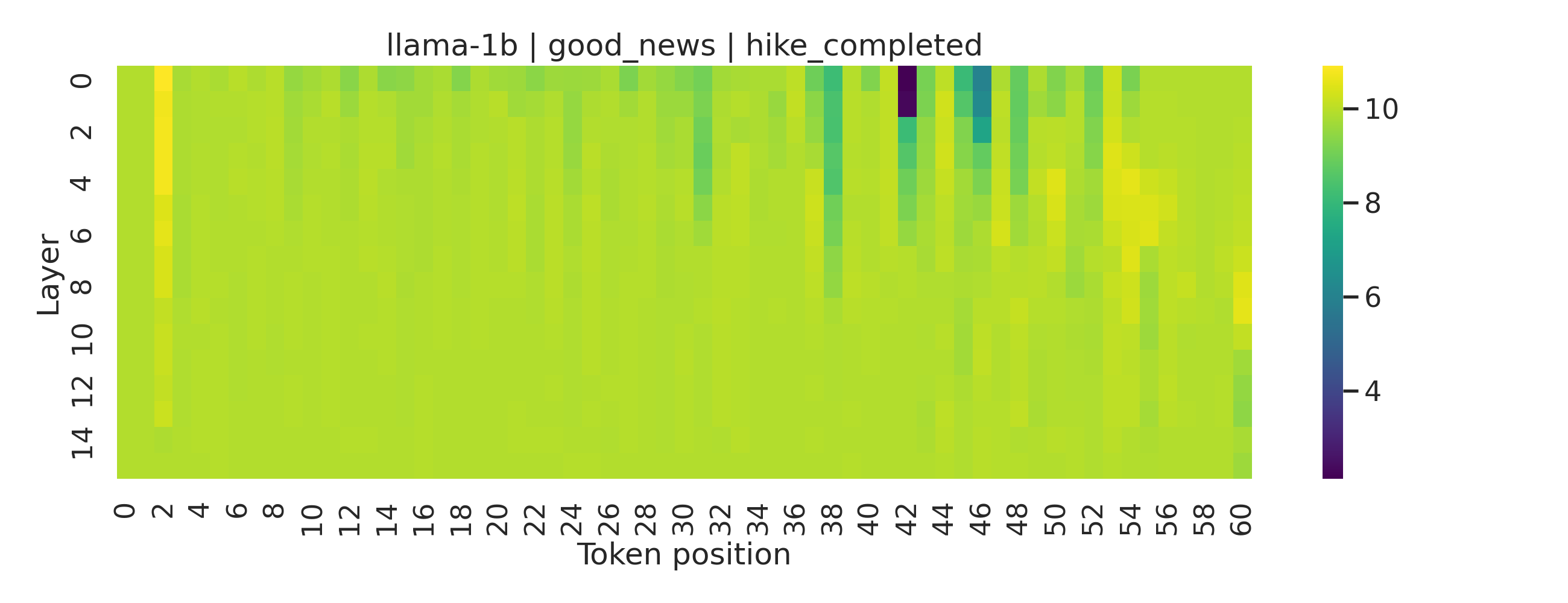}
\end{minipage}
\hfill
\begin{minipage}{0.48\linewidth}
  \centering
  \includegraphics[width=\linewidth, trim=0 0 0 36.5pt, clip]{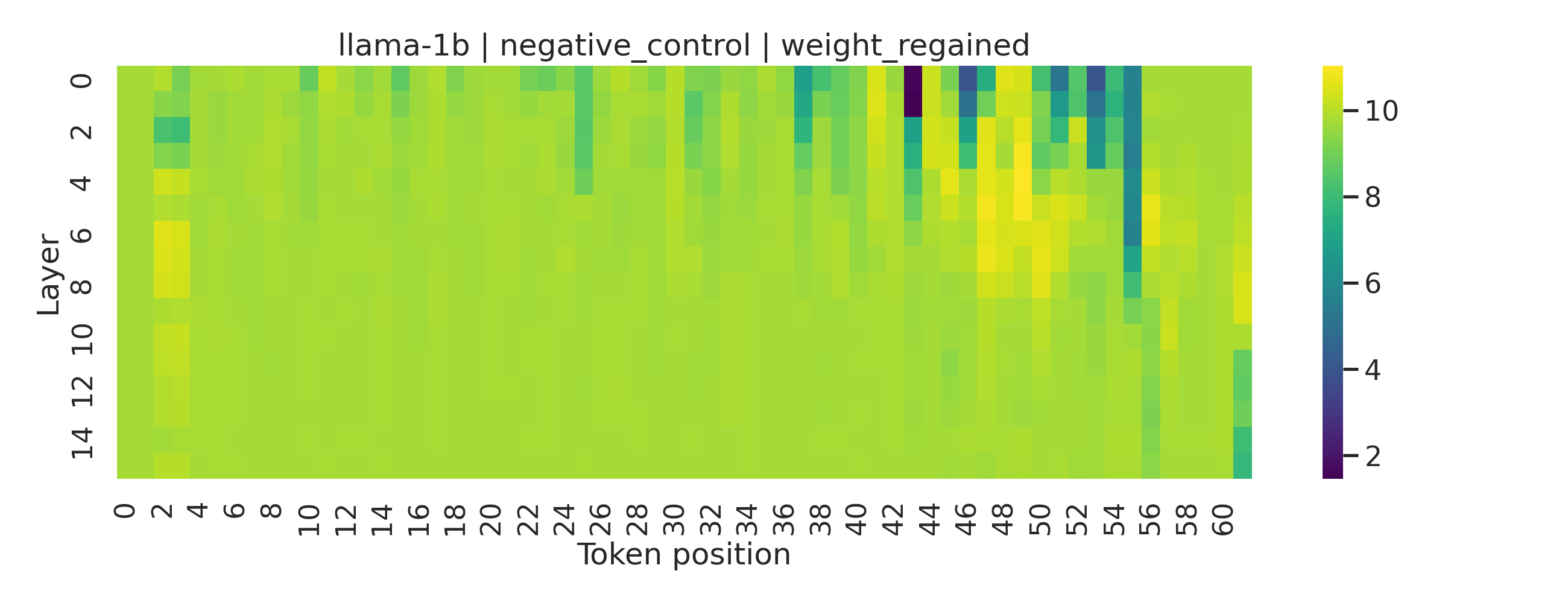}
\end{minipage}
\caption{Llama-1B residual stream patch heatmaps. The negative-control
  heatmap (right) shows activity concentrated in layers 0--3 while
  the good-news heatmap (left) shows activity spread across layers
  8--15. The same early-late asymmetry holds as in Qwen-3B despite
  the architectural difference.}
\label{fig:heatmaps_llama}
\end{figure}

\end{document}